\documentclass{article}

\usepackage{arxiv}

\usepackage[utf8]{inputenc} 
\usepackage[T1]{fontenc}    
\usepackage{hyperref}       
\usepackage{url}            
\usepackage{booktabs}       
\usepackage{amsmath}
\usepackage{amsfonts}       
\usepackage{nicefrac}       
\usepackage{microtype}      
\usepackage{lipsum}

\usepackage{float}
\usepackage{array}
\usepackage{graphicx}
\usepackage{subfig, caption}
\usepackage{multirow}
\usepackage{tabularx}
\usepackage{cleveref}

\usepackage{comment}

\title{Towards Learning \\ Cross-Modal Perception-Trace Models}

\author{
  Achim Rettinger\\
  Trier University\\
  Trier, Germany \\
  \texttt{rettinger@uni-trier.de} \\
   \And
 Viktoria Bogdanova\\
  Karlsruhe Institute of Technology\\
  Karlsruhe, Germany\\
  \texttt{uzdag@student.kit.edu} \\
   \AND
  Philipp Niemann \\
  National Institute for Science Communication \\
  Karlsruhe, Germany \\
   \texttt{niemann@nawik.de} \\
}

\begin{document}
\maketitle

\begin{abstract}
  Representation learning is a key element of state-of-the-art deep learning approaches. It enables to transform raw data into structured vector space embeddings. Such embeddings are able to capture the distributional semantics of their context, e.g.\ by word windows on natural language sentences, graph walks on knowledge graphs or convolutions on images. So far, this context is manually defined, resulting in heuristics which are solely optimized for computational performance on certain tasks like link-prediction. However, such heuristic models of context are fundamentally different to how humans capture information. For instance, when reading a multi-modal webpage (i) humans do not perceive all parts of a document equally: Some words and parts of images are skipped, others are revisited several times which makes the perception trace highly non-sequential; (ii) humans construct meaning from a document's content by shifting their attention between text and image, among other things, guided by layout and design elements. 
  In this paper we empirically investigate the difference between human perception and context heuristics of basic embedding models. We conduct eye tracking experiments to capture the underlying characteristics of human perception of media documents containing a mixture of text and images. Based on that, we devise a prototypical computational perception-trace model, called CMPM. We evaluate empirically how CMPM can improve a basic skip-gram embedding approach. Our results suggest, that even with a basic human-inspired computational perception model, there is a huge potential for improving embeddings since such a model does inherently capture multiple modalities, as well as layout and design elements.
\end{abstract}


\maketitle


\section{Introduction}

\begin{figure}
  \includegraphics[width=\textwidth]{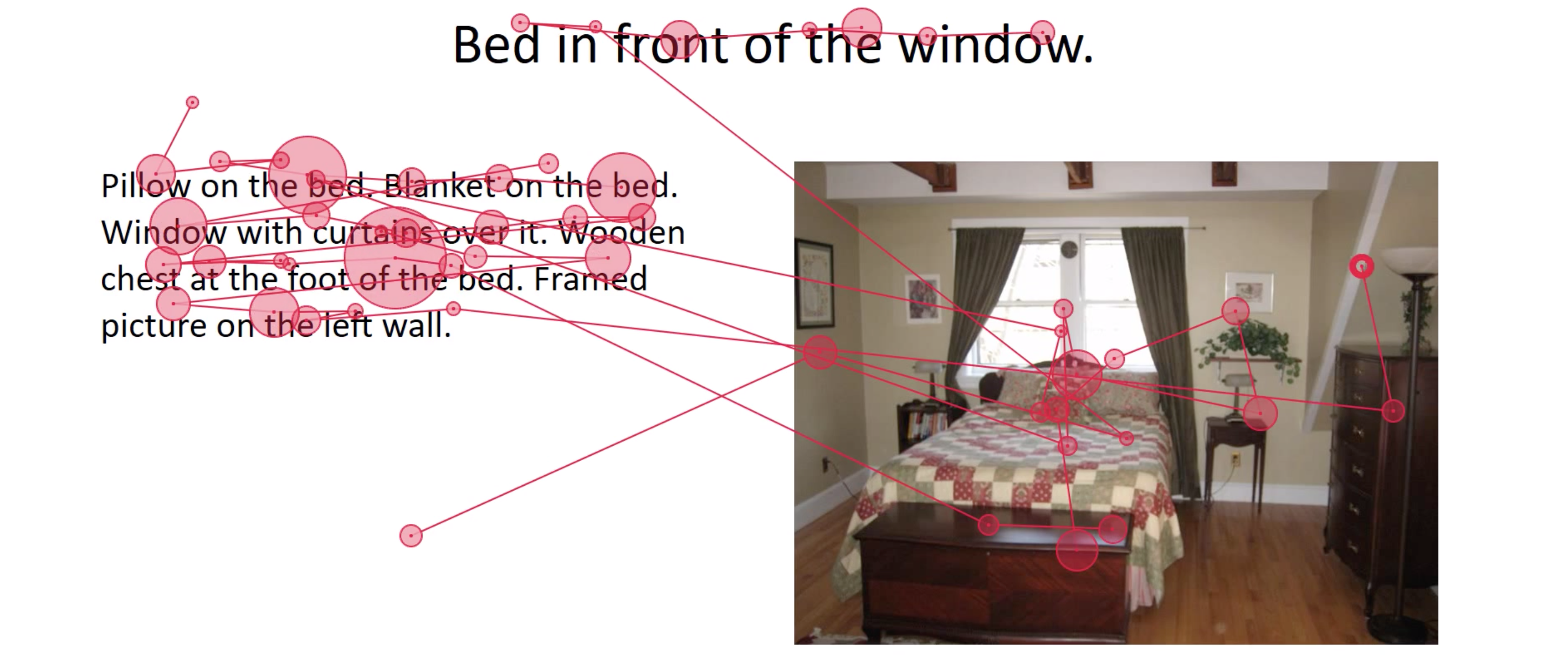}
  \caption{Perception Trace obtained by eye tracking.}
  \label{fig:eye-track}
\end{figure}

Representation learning techniques are key to the machine learning revolution of the last years. For instance, in Natural Language Processing they encode the context of a word based on the surrounding words in the sentence, in Computer Vision they capture visual traits from convolutions of image patches and in Network Analysis they encode the network structure of a node's neighborhood. Such learned vector representations, also called embeddings, have shown great performance in many benchmark tasks and real-world applications.

One further, very recent, advancement was achieved by Attention Mechanisms \cite{vaswani2017attention} which learn to capture the relevance of elements of the input (e.g. words in a sentence) for contextualizing other elements of the input. 
In case of natural language processing the latest successes of transformer-based models like Bert \cite{devlin2018bert} and XLNet \cite{yang2019xlnet} demonstrate that performance can be significantly increased, if the attention is calculated over the full input and not only for the last elements (e.g. the full sentence and not only the previous word) compared to Recurrent Neural Network approaches like LSTMs \cite{hochreiter1997long}. Such findings are intuitive, since typically not all parts of a sentence have the same impact on the overall meaning and certain elements need to be combined to obtain a deeper meaning. 

The role of human perception in the extraction of meaning from documents has been investigated in depth in scientific disciplines beyond artificial intelligence. One instance is media studies, where results of eye-tracking experiments suggest that oversimplifying assumptions are being made by all state-of-the-art embedding algorithms (see Section \ref{Sec:relatedwork}). Typically trained on large text corpora, large collections of images or large graph-structured data sources they ignore essential cues that are key to human perception. 

Aspects of layout and design are ignored in any previous work, as are different dependencies of modalities in a multimedia document: Elements expressed in different modalities (e.g., objects in images and words in texts) are intertwined: Humans construct meaning from a document's content by moving their attention between text and image as, among other things, guided by layout and design elements. 
Not all parts of the document are perceived equally: Some words are skipped, others are revisited several times. So far, machine reading of content is typically done sequentially as required by the raw document encodings (e.g., word by word and sentence by sentence) ignoring the interaction with other modalities.

In this work we propose the hypothesis that context models which are inspired by human-perception have advantages over heuristically defined context models. The rationale being, that media documents are designed by humans for humans, not for machines and hence are best  perceived in a human-like way.

To test our hypothesis we investigate human perception of multi-modal documents via eye tracking and derive a simple computational perception-trace model from our findings, called CMPM. When plugging this model into a basic embedding method like skip-gram \cite{mikolov2013efficient} our results indicate that the performance can greatly benefit. This is likely due to 
a) the capturing of the interdependencies of information appearing in multiple modalities in the document; 
b) exploiting the patterns of how humans perceive and interact with multimedia documents through the integration of layout elements.

\section{Investigating Human Perception Traces with Visual Genome Data}

To test the hypothesis that human-inspired Cross-Modal Perception-trace Models (CMPM) have advantages, we first need to investigate how humans perceive multimedia documents in order to derive a computational model. We do so, by conducting eye tracking studies on the perception of basic documents, consisting of a heading, an image and a short description of the image. 

In order to being able to learn embeddings across modalities, the mentioned and depicted entities in those documents need to be annotated and aligned. We leverage the Visual Genome Data Set (VG) \cite{krishnavisualgenome} for that, a widely used multi-modal data collection\footnote{The data set contains 108,077 real-world images selected from the intersection of the MS COCO \cite{lin2014microsoft} and YFCC100M \cite{thomee2015yfcc100m}}. The visual content in VG is densely annotated and the image scene is described through multiple human-generated descriptions of image regions localized by bounding boxes. Each image has an average of 50 regions. The region descriptions vary in their length as well as semantic properties, comprising short noun phrases or full sentences. Multiple descriptions can be provided to the regions referring to the same object, increasing the redundancy in annotations. In addition, VG provides well structured information about the depicted objects and their relationships which enables the representation of each image in the form of a cross-modal scene graph. 

In order to make eye tracking experiments feasible the redundancy in the initial region descriptions needs to be eliminated. This step is inspired by the salience of the words in a sentence \cite{erkan2004lexrank}. To identify the key elements of a scene and create its compressed description we construct a graph-based document representation in form of an undirected unweighted graph for each image plus all descriptions. The vertices of generated graphs are equivalent to the image objects, whereas the edges correspond to their relationships to each other. 

\begin{figure*} 
  \includegraphics[width=\textwidth]{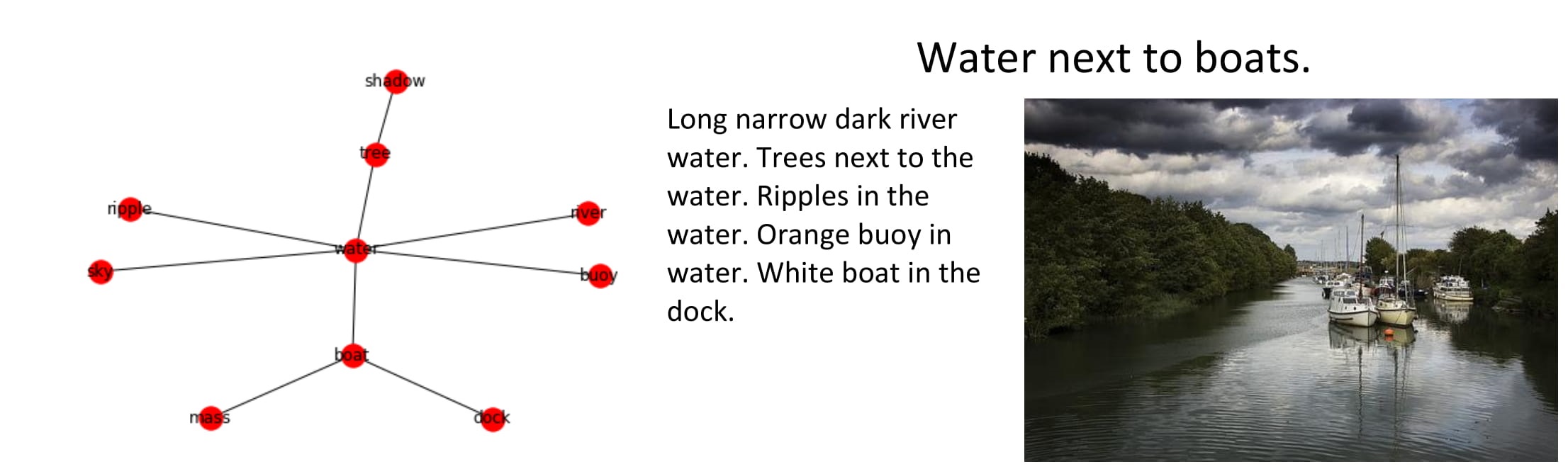}
  \caption{Multimedia document (headline + image + description) generated from the Visual Genome data set (right) and the corresponding scene graph (left).}
  \label{fig:scene-graph}
\end{figure*}

Next, we add layout elements to each image by constructing a headline and a short textual description. 
In accordance with the findings of \cite{matas2017comparing}, we chose degree centrality, i.e. the number of direct neighbours of a graph node, as a salience measure. Extracting a pair of linked nodes with the highest summed number of neighbours, i.e.,\ the largest total degree, yielded the headline and captions for our experimental data. 
By applying a graph-based ranking approach we obtained for each image a brief textual summary containing three to five sentences from all captions provided. For each of the selected entities we, thus, generated a 3-tuple consisting of a headline, description and corresponding image. An example of a built scene graph and extracted experiment data is shown in Figure~\ref{fig:scene-graph}. The original spelling was retained, introducing noise in the generated data. 

\begin{figure*}[t]
\centering
\subfloat[Subjects' eye behavior represented as sequence chart.]{\label{fig:chart}\includegraphics[width=0.5\textwidth,keepaspectratio=True]{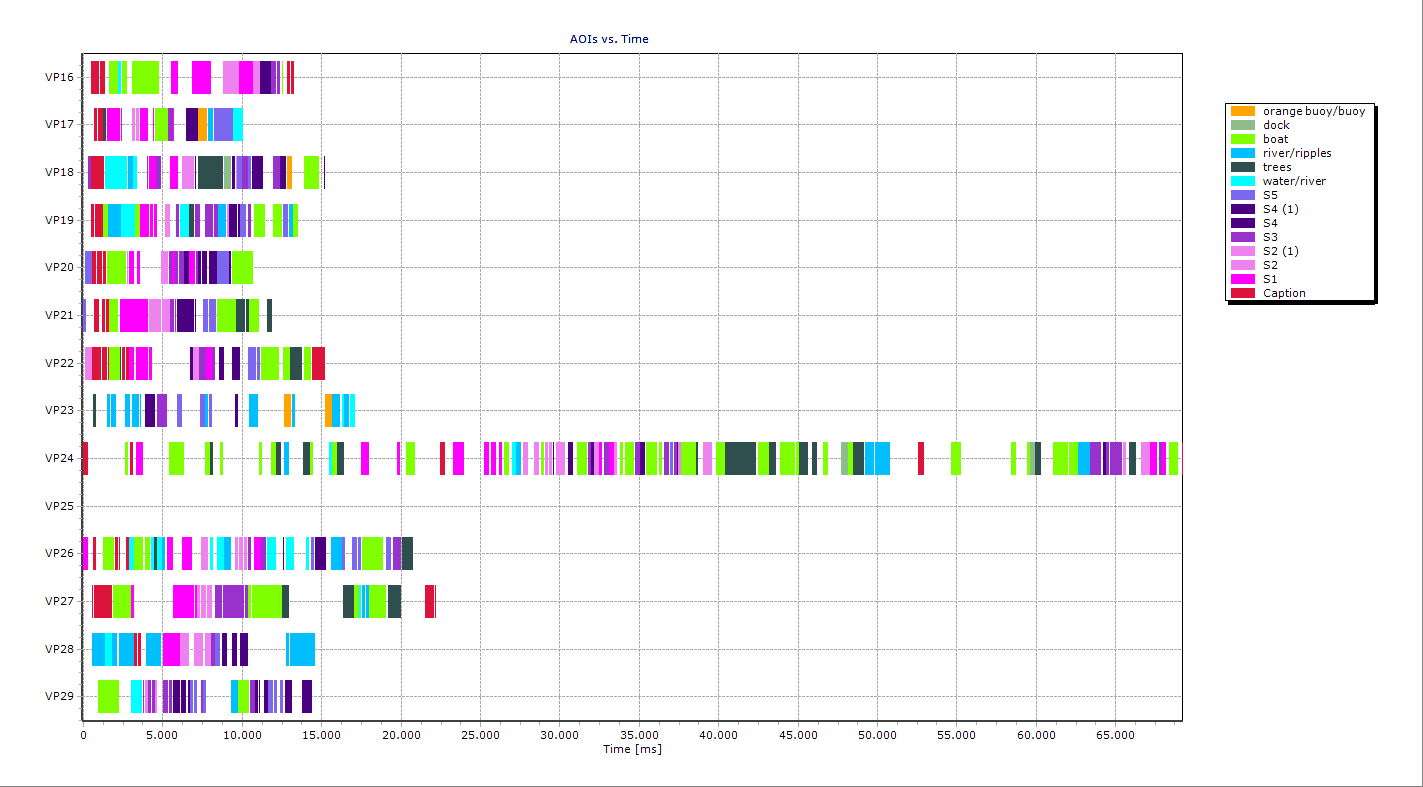}}
\subfloat[Highlighted areas of interest.]{\label{fig:aoi}\includegraphics[width=0.5\textwidth,keepaspectratio=True]{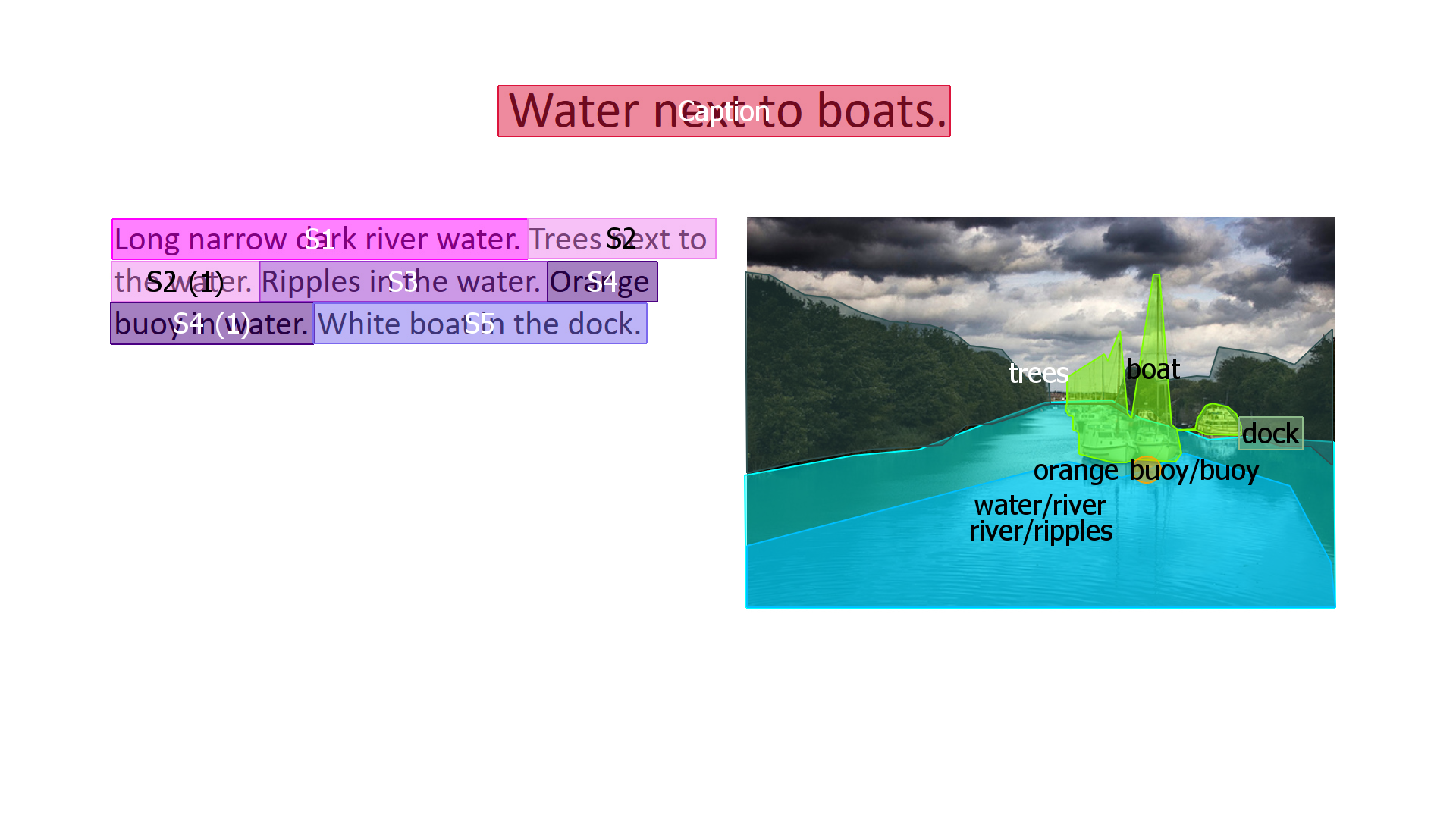}}
\caption{Example documents constructed for eye-tracking study and their evaluation.}
\label{fig:e_tr}
\end{figure*}

Finally, to extract a model of how humans perceive concepts presented in a form of multimedia documents a reception study involving eye-tracking experiments with 28 participants was conducted. 
The subjects were asked to view a pre-selected sequence of presentation slides each depicting a 3-tuples from the VG data.
Each of them viewed a presentation consisting of 16 randomly arranged slides. We tested two layout types for each entity: The image was placed either left or right from the related text. A slide generated for one of the chosen entities in one of the layouts was shown to 14 different subjects. The eye-tracking equipment was applied to register the gaze trajectory and leaps between text and image elements of each test person. We obtained individually varying results with regard to the order as well as the duration of viewing the respective document component (see Figure~\ref{fig:chart}). All recorded perception traces are made available online\footnote{\url{https://drive.google.com/open?id=1s7zJZAMVZYvA034A4waVPuIzGTxf65rj}}.

The conducted measurements revealed a frequently repeating pattern in the perception of multimedia data. Beginning with the entities in the headline, as they appear in the text, test people often switched their visual attention to the corresponding image regions depicting the named objects. Textual information in each following sentence in the summary was processed in a similar manner as in the headline, whereas image data delimited by bounding boxes was viewed after reading the respective text section if a new object not named in previous sentences was mentioned. 

\begin{figure}
    \centering
  \includegraphics[width=.5\textwidth]{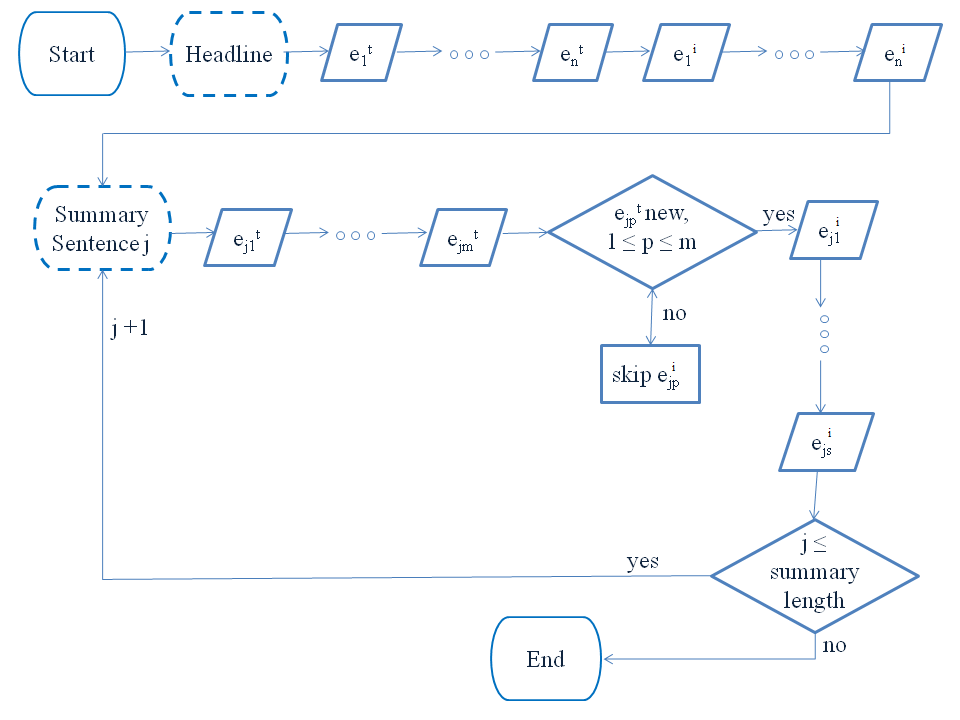}
  \caption{Exemplary flow chart for the perception trace extraction with $e^{t}$ and $e^{i}$ being text and visual entities, respectively; n, m, s - number of text or visual entities in a caption or the j-th summary sentence with $s \leq m$, $3 \leq j \leq 5$.}
  \label{fig:flow-chart}
\end{figure}

Given this pattern we then could derive a computational perception model, by manual inspection. When parsing a multi-modal document in the above described manner a linear sequence of text and image entities can be obtained, hereafter called perception trace. The procedure is visualized in flow chart in Figure~\ref{fig:flow-chart} (the notation is introduced in Section~\ref{Sec:CMPM}. While this still is an obvious oversimplification we argue that it comes closer to human perception than the common sequential word-by-word parsing of text that is independent of the visual information.

\section{Cross-modal Perception-trace Skip-gram Model}
\label{Sec:CMPM}


Given the computational model of extracting perception traces from multi-modal documents as described in the previous section we can now try to capture this information in embedding vectors $v$. Hereby, the captured entities $e$ can occur in different modalities $m$, i.e. images $i$ or text $t$. We denote $v_{e^m}$ as the vector representation of entity $e$ as occurring in modality $m$.

As an input, we start with $|m|$ independent matrices. With the two modalities, text and image, this is matrix $V_{t}^{\text{pre-trained}}$ with stacked pre-trained word embeddings and matrix $V_{i}^{\text{pre-trained}}$ containing all pre-trained image representations. The dimensions of the input matrices is the number of distinct entities occurring in perception traces in both modalities\footnote{Which is identical, as entities are annotated and aligned in the VG data across modalities.}, $\times$ the dimensionality of the latent space, as output of the pre-trained word and image embedding method\footnote{Which is different, since pre-trained image embeddings typically require more dimensions than word embeddings.}.

Three transformation steps of the vector spaces $V^{\text{pre-trained}}$ are needed to achieve a common cross-modal embedding space that captures the perception trace information:

\begin{enumerate}
	\item Reduce $V^{\text{pre-trained}}$ for each $m$ to a vector space with common dimensions $V^{\text{reduced}}$. This can be achieved with any unsupervised dimensionality reduction technique.
	\item Optimize the embeddings $v \in V^{\text{reduced}}$ for each $m$ independently according to their cross-modal distributional semantics (see Equation \ref{Equ:softmax}).
	\item Combine the optimized embeddings $v \in V^{\text{reduced}}$ of each $m$ into one cross-modal joint space $V^{\text{joint}}_x$. Trivially, this is done by element-wise addition (as proposed in \cite{pennington2014glove}) across all modalities: 
	\begin{equation*}
	    v_{e^x} = \sum_{s=1}^{|m|} v_{e^{s}}
	\end{equation*} 
\end{enumerate}

In this paper we intend to stick to established techniques like word2vec as close as possible. Our aim is not to propose a new embedding method, but to assess the impact of perception trace information for standard embedding techniques. We picked word2vec, since it deploys an established simple word-window-based context model. 

Accordingly, the training objective in Step~2 of our model is to maximize the classification of a output entity $e_o$ given a center entity $e_c$. The context is hereby defined by the perception trace instead of a word-window and the training objective can be expressed as maximizing the soft-max probability:

\begin{equation}
p(e_{o} | e_{c}) = \cfrac{exp(v_{e_o^m}^T v_{e_c^m})}{\sum_{j=1}^{|e|} exp(v_{e_j^m}^T v_{e_c^m})}
\label{Equ:softmax}
\end{equation}

Please note, that any combination of modalities for $e_o$ and $e_c$ is possible: a text-entity can be followed by a text-entity or image-entity and vice versa.

\section{Empirical Evaluation}

\subsection{Implementation}

For the training set we replaced text and image units in each perception trace with pre-trained dense vector representations. To obtain word embeddings we exploited word2vec\footnote{\url{https://code.google.com/archive/p/word2vec/}} \cite{mikolov2013efficient} and to compute vector representations for visual entities in the generated perception traces we chose the Inception-V3 model\footnote{\url{http://download.tensorflow.org/models/inception_v3_2016_08_28.tar.gz}} \cite{szegedy2016rethinking}
The resulting word and image vectors have 300 and 2048 dimensions, respectively. 

Since our prime target is to evaluate embeddings based on established benchmark data sets we only select images from the VG data set that contain objects that occur in the intersection with the vocabulary of merged evaluation sets - MEN \cite{bruni2014multimodal}, SimLex-999 \cite{hill2015simlex}, MTurk-771 \cite{halawi2012large} and WordSim-353 \cite{finkelstein2002placing}. The resulting list contains 1702 words. We focus on the frequently used entities mentioned in VG at least 1000 times in total and five times per image description.

Our final training data comprises 55,237 documents, each consisting of a variable length sequence of pre-trained vectors of different modalities, each vector being either a word embedding or an image embedding. The training data is made available online\footnote{\url{https://drive.google.com/open?id=1s7zJZAMVZYvA034A4waVPuIzGTxf65rj}}. 

To transform the two input matrices $V^{\text{pre-trained}}_{t} \in \mathbb{R}^{5470 \times 300}$ and $V^{\text{pre-trained}}_{i} \in \mathbb{R}^{5470 \times 2048}$ to the same dimensions we used a standard autoencoder with one fully-connected hidden layer containing 100 neurons. 
The values for the initialization weight matrices $w_{1} \in \mathbb{R}^{300 \times 100}$ and $w_{2} \in \mathbb{R}^{2048 \times 100}$ were generated randomly, following standard normal distribution truncated to the range $[-0.1, 0.1]$. $tanh$, was used as an activation function and the mean squared error as the loss function. The learning was performed using the Adam opitimizer as described in \cite{kingma2014adam}. After training the model for five epochs we extracted 100-dimensional distributed text and visual representations and stacked them to a new matrix $V^{\text{reduced}} \in \mathbb{R}^{10940 \times 100}$, building the combined text and visual vocabulary.
To determine the context of each perception trace element we used a window size of five, treating five entities from history and five entities from the future of the target element as positive samples. 

Starting with a random pair consisting of a target and context entity represented through a one-hot encodings, we multiplied each pair element with the weight matrix $M$ to yield the matching embeddings of the given entities (see Step~1 in Section~\ref{Sec:CMPM}). Then we estimate the softmax (Step~2) and optimize the error using
RMSprop, an adaptive gradient descent algorithm introduced in \cite{tieleman2012lecture}. We applied the learning rate $\alpha = 0.001$

After training the model for ten epochs we extracted two weight matrices containing representations of the target and the context entities, respectively. 
To yield the final entity embeddings we then calculated the component-wise sum of the learned matrices (Step~3). The evaluation of the embedding quality is described in the following subsection.

\subsection{Semantic Similarity}

As the baseline for the following tests we used the pre-trained 300-dimensional word2vec embeddings as well as 100-dimensional GloVe\footnote{\url{https://nlp.stanford.edu/projects/glove/}} \cite{pennington2014glove} word vectors trained on the combination of Gigaword5 and Wikipedia2014, corpora containing six billion tokens. 
To test the quality of the embeddings obtained by training our Cross-Modal Perception-trace Model we tested whether our embeddings are able to capture human notion of word similarity and relatedness. We compared subjective human scores of semantic similarity of word pairs accumulated in data sets MEN, WS-353 and SimLex-999 with the cosine distance between the corresponding entity vectors. Since we can learn embeddings only for the intersection of entities of those data sets with VG data, we end up with subsets of size 38
of the mentioned evaluation data sets. Thus, the numbers stated below are not directly comparable to the results of other models evaluated on the complete data collection.  

\begin{table}
\begin{center}
\begin{tabular}{r|c}  
  \textbf{Embeddings} & \textbf{Weighted average} \\
  \hline
  \textbf{CMPM} & \textbf{0.755} \\
  \textbf{GloVe} & 0.451 \\  
  \textbf{word2vec} & 0.565 \\
\end{tabular}
\end{center}
\caption{Spearman's rank correlation averaged over all evaluation data sets.}
\label{Tab:rho}
\end{table}

We computed the Spearman's average rank correlation over all evaluation data sets weighted according to their respective size and obtained the score of 0.451 and 0.565 for GloVe and word2vec representations, respectively. Our embeddings seems to match the human notion of similarity considerably better, by achieving the rank correlation of 0.755 (see Table~\ref{Tab:rho}). All embeddings show comparable performance on the subsets of MEN and WS-353, word2vec representations being less correlated with the scores provided in the WS-353 subset. The difference in the weighted results lies mainly in the worse performance of the word2vec and GloVe vectors on the subset of SimLex-999. A possible explanation of this gap is that the evaluation data sets MEN and WS-353 reflect human judgments of the relatedness of the word pairs, SimLex-999 explicitly measures concept similarity \cite{bruni2014multimodal, finkelstein2002placing, hill2015simlex}. Although relatedness and similarity are highly connected concepts, strongly associated entities are not necessarily similar. 
More precisely the pairwise correlation between the ratings in the SimLex-999 subset and the GloVe vectors is negative, word2vec and our model score on the Spearson's rank correlation 0.205 and 0.462, respectively. 


Apart from the semantic similarity of word pairs, SimLex-999 provides information on other properties of words such as their concreteness score. The selected entities are directly perceptible, tangible concepts with scores in the range $[4.46, 5]$ on a scale from 1 to 7. Thus, a better performance of our perception model on the similarity task can be explained through the integration of the visual context which provides complementary information for concrete entities not available to distributed models relying only on one modality.

\subsection{Dimensionality Reduction}

We projected the learned vectors down to three dimensions using Principal Component Analysis \cite{hotelling1933analysis}.  
While the obtained principal components (PC) of the GloVe and word2vec representations are of a mainly mixed nature and cannot be easily interpreted, our model allows clearer definition of the found feature subspaces. We observed that the first component is strongly associated with nature and recreational activities, the second component relates to the household and domestic life, the third component is correlated with entities occurring in the road traffic context. Thus, it can be stated that our embeddings are able to capture semantically meaningful information about the selected entities and their relationships to each other (cmp.~Table~\ref{pc}).

\begin{table}
\centering

\label{pc}
\begin{tabular}{m{0.37cm}|m{7.65cm}}
\hline 
\textbf{PC} & \centering \textbf{Entities} \tabularnewline
\hline

  \multicolumn{2}{c}{\textbf{\textit{Our Embeddings}}} \\
  \hline
  PC1 & beach, bear, bench, bird, boat, building, cow, elephant, field, giraffe, grass, horse, skateboard, snow, tree, water \\
  PC2 & bed, cat, dog, girl, glass, man, pizza, plate, shirt, table, wall, woman \\
  PC3 & bus, car, motorcycle, road, sign, street, toilet, train, truck, window \\
  \hline
  \multicolumn{2}{c}{\textbf{\textit{GloVe}}} \\
  \hline
  PC1 & beach, bear, bird, cow, field, grass, road, snow, tree, water \\
  PC2 & bed, bench, building, bus, car, glass, plate, shirt, sign, street, table, toilet, train, truck, wall, window \\
  PC3 & boat, cat, dog, elephant, giraffe, girl, horse, man, motorcycle, pizza, skateboard, woman \\
  \hline
  \multicolumn{2}{c}{\textbf{\textit{word2vec}}} \\
  \hline
  PC1 & beach, bench, building, field, grass, plate, road, snow, street, tree, wall, water \\
  PC2 & bear, bed, bird, cat, cow, dog, elephant, giraffe, girl, glass, horse, man, sign, table, toilet, window, woman \\
  PC3 & boat, bus, car, motorcycle, pizza, shirt, skateboard, train, truck \\
  \hline
\end{tabular}
\caption{Principal component analysis results.} 
\end{table}

Three principal components of the GloVe, word2vec and vectors learned by applying the Cross-Modal Perception-trace Model (CMPM), explain about $32.63\%$, $24.44\%$ and $65.03\%$ of the data variance, respectively. The first five principal components are sufficient to describe about $80\%$ of the variance in our embeddings, whereas to achieve comparable results using the GloVe vectors, having the same dimensionality as ours, and 300-dimensional word2vec representations we need 14 and 18 components, respectively. This leads to the assumption that with our model we can encode most of the essential properties of the data using less than 100 vector features and achieve comparable performance as models making use of higher dimensional embeddings while saving storage space and computational time.

\subsection{Concept Categorisation}

We proceeded by testing the quality of the learned embeddings on the task of the entity categorization. We used graph- and density-based clustering algorithms not requiring the number of clusters as an input parameter. To visualize the obtained clusters we applied t-Distributed Stochastic Neighbor Embedding (t-SNE) \cite{maaten2008visualizing}.

\begin{figure*}
\centering
\subfloat[Our embeddings]{\label{ap_embed}\includegraphics[width=0.33\textwidth]{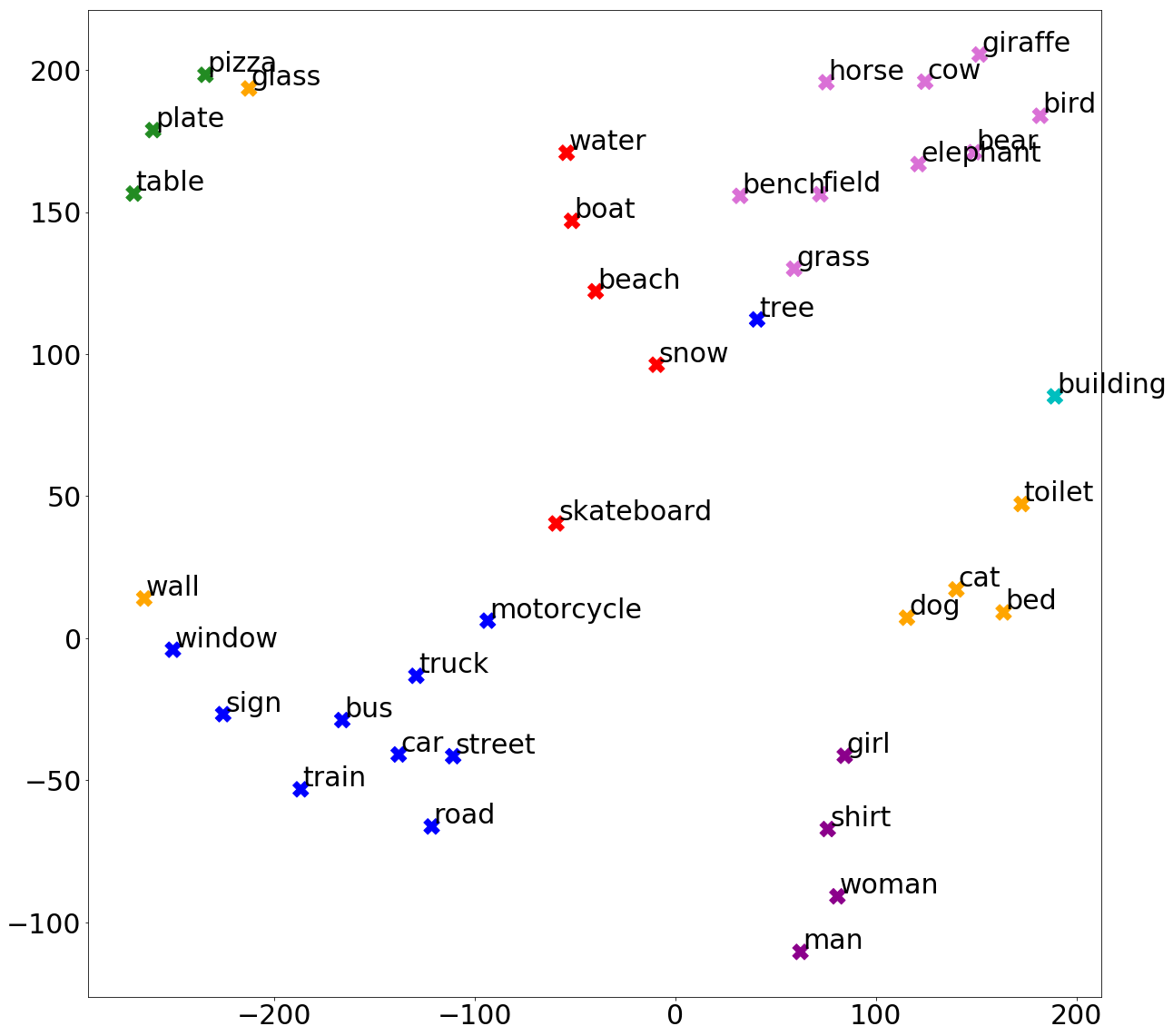}} \subfloat[GloVe]{\label{ap_glove}\includegraphics[width=0.33\textwidth]{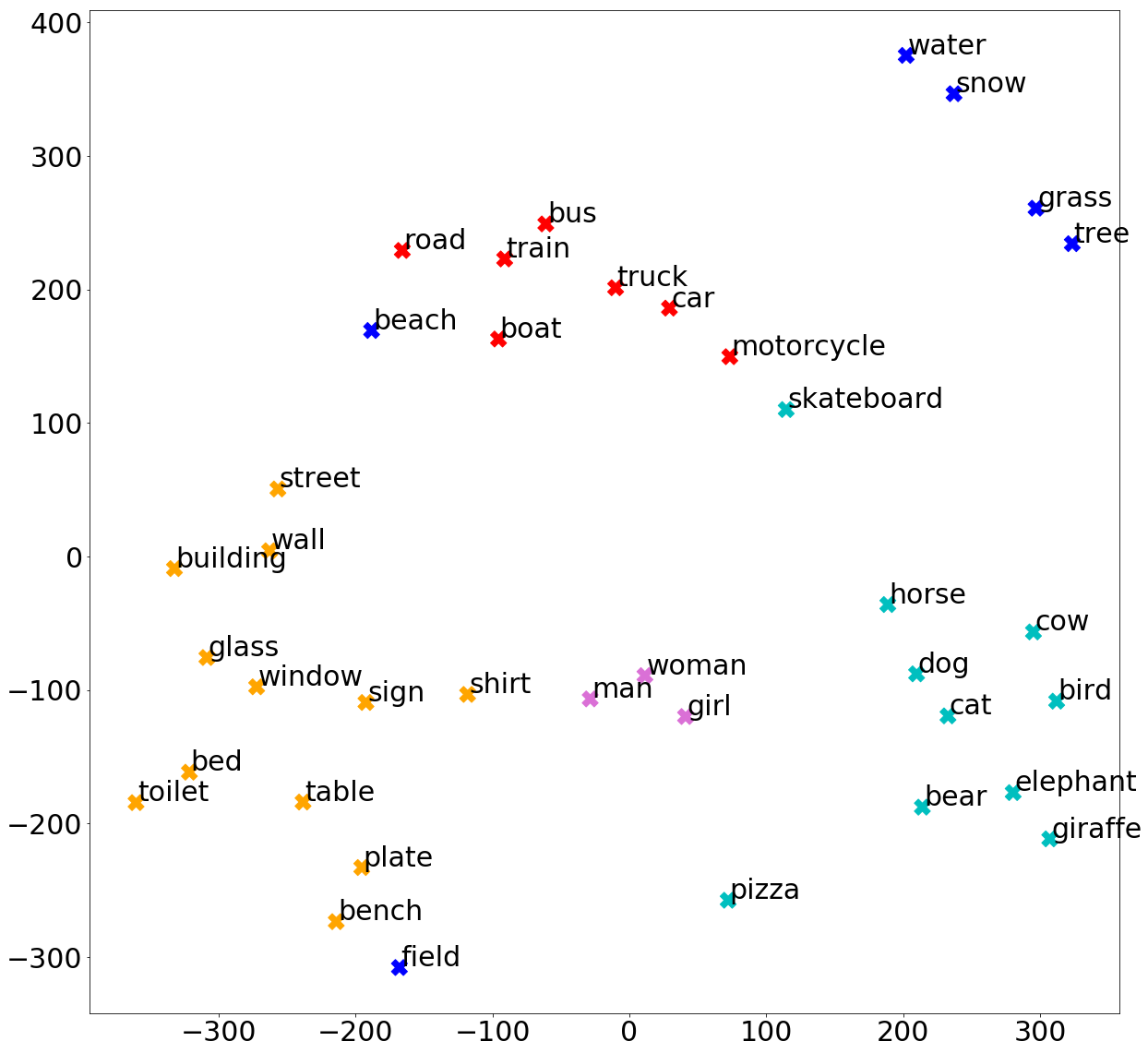}}
\subfloat[word2vec]{\label{ap_w2v}\includegraphics[width=0.33\textwidth]{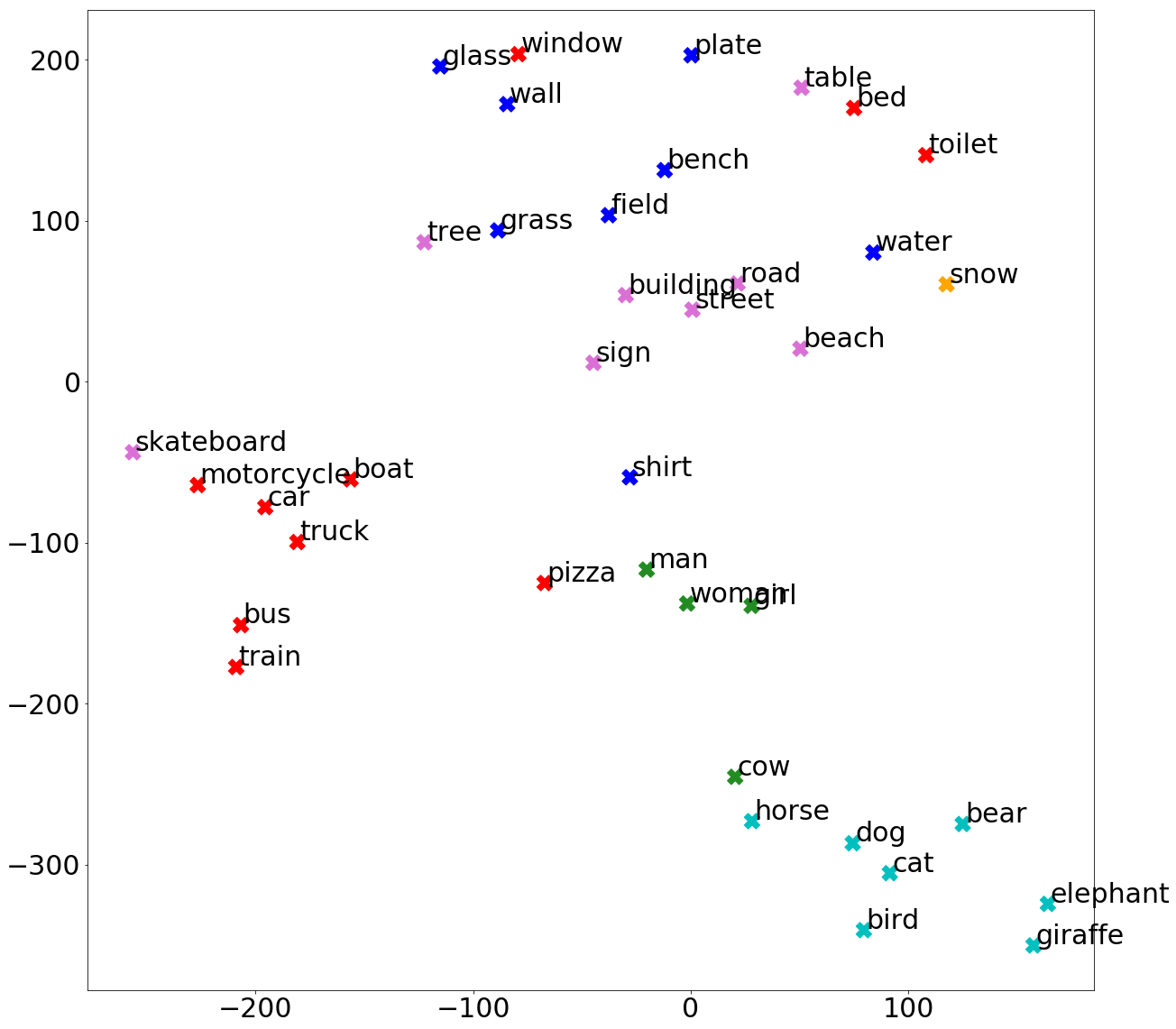}} \\
\subfloat[Our embeddings]{\label{hdbscan_embed}\includegraphics[width=0.33\textwidth]{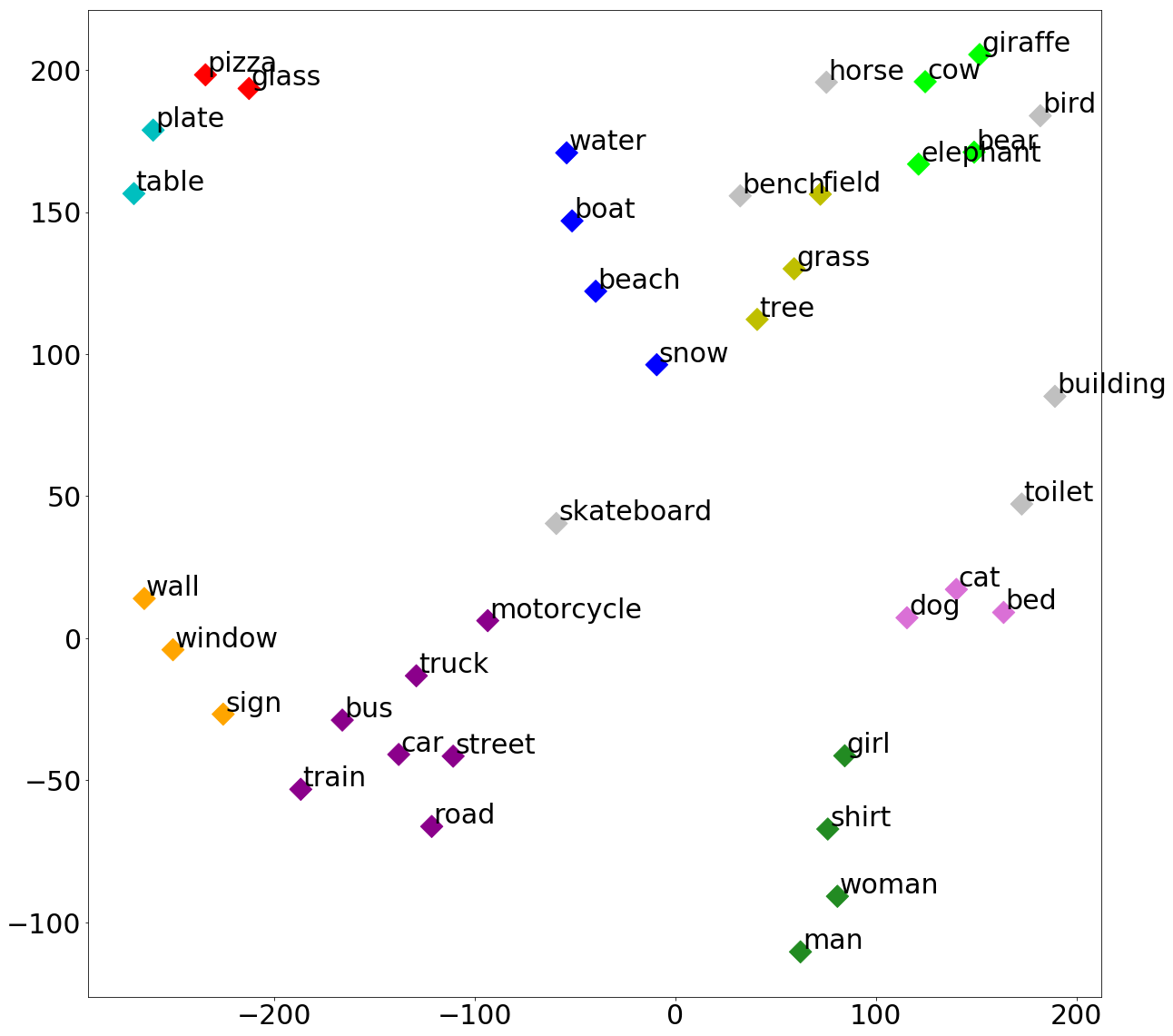}} 
\subfloat[GloVe]{\label{hdbscan_glove}\includegraphics[width=0.33\textwidth]{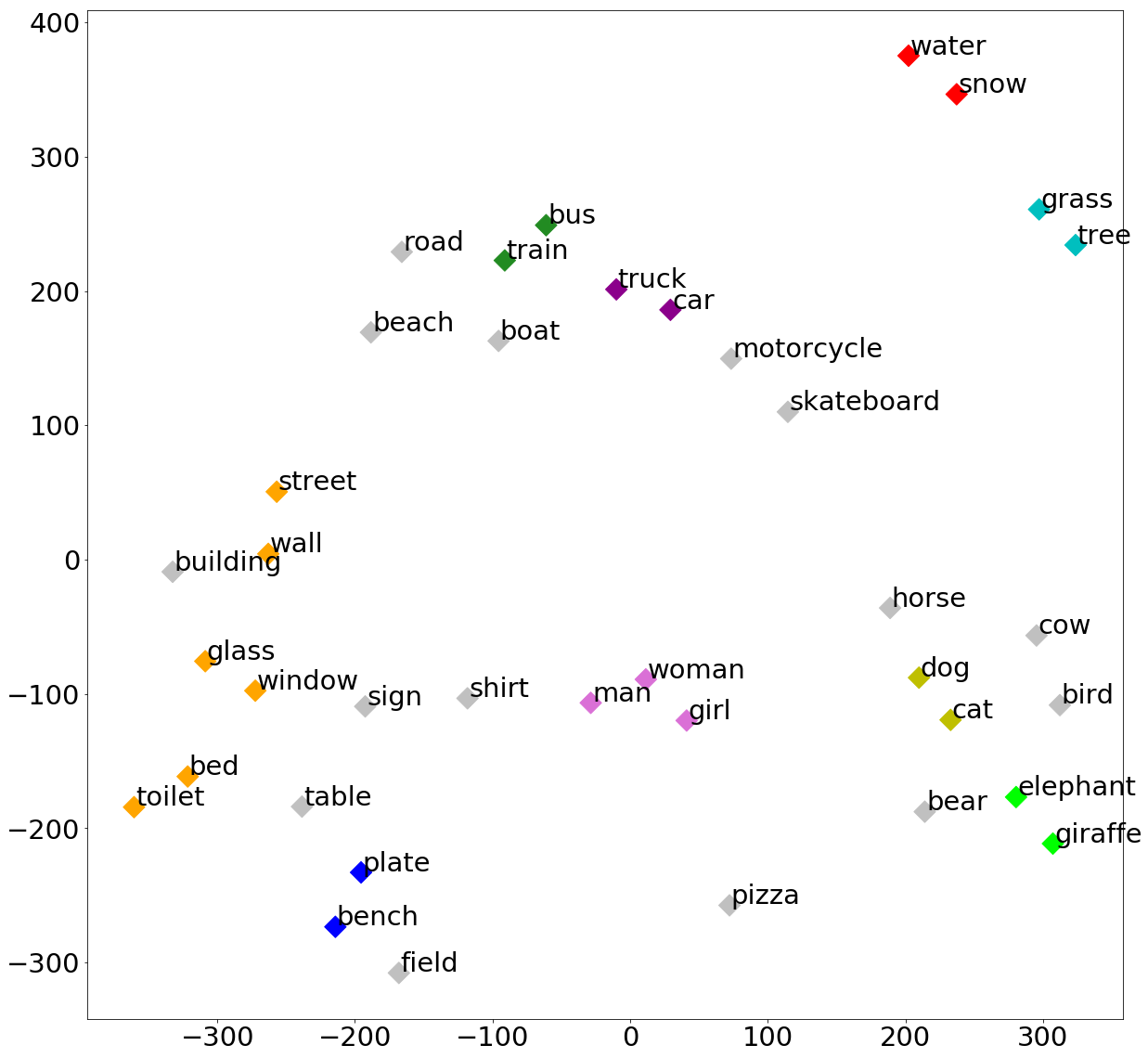}}
\subfloat[word2vec]{\label{hdbscan_w2v}\includegraphics[width=0.33\textwidth]{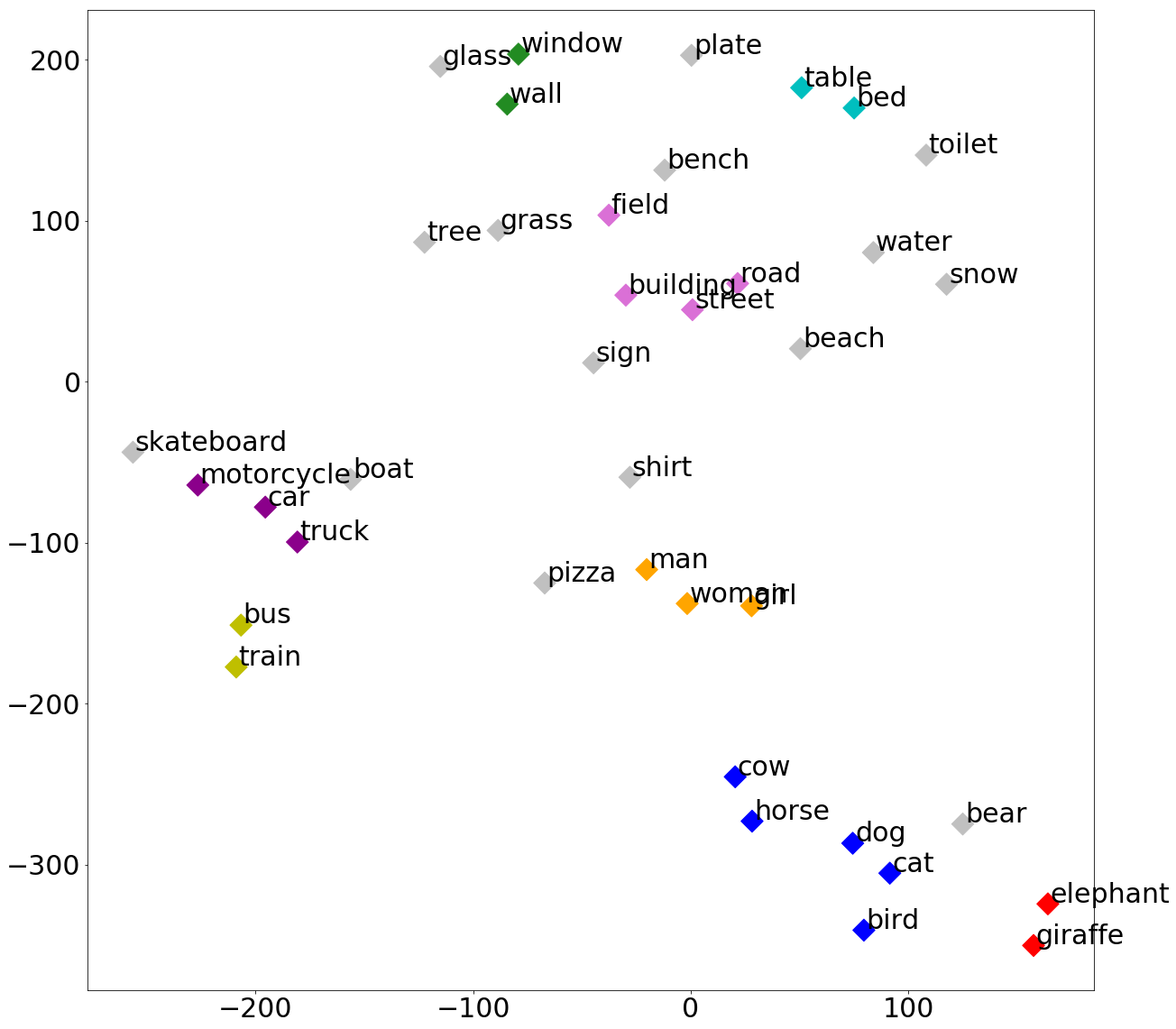}} 
\caption{Results of the applied AP clustering (top row) and HDBSCAN (bottom row).}
\label{fig:cluster}
\end{figure*}

To categorize our data samples we exploited the Affinity Propagation (AP) clustering \cite{frey2007clustering} and used the median of the samples similarities as input preferences and the damping factor of $0.75$. The plotted data makes evident that the clusters obtained for the word2vec embeddings are noisier than other compared representations with five of overall six identified clusters containing unrelated objects. (see Figure~\ref{ap_w2v}). 
The algorithm yielded for the GloVe and our vectors five and seven clusters, respectively. (see Figures \ref{ap_embed} and \ref{ap_glove}). 
Although some entities are clearly out of context the obtained clusters are mostly consistent and have distinguishable identity.

For further evaluation we use the HDBSCAN \cite{campello2013density} - Hierarchical Density-Based Spatial Clustering of Applications with Noise. By setting a minimal number of objects in a cluster $m_{pts}$ to $2$ we obtain reasonable results for all compared representations, related entities being assigned to one cluster. 
It is noticeable that word2vec and GloVe vectors mostly build smaller clusters containing two or three objects, whereas nearly $40\%$ of the entities are categorized as noise. (in Figures \ref{hdbscan_embed}-\ref{hdbscan_w2v} labelled in grey). 
Applying the algorithm to our embeddings resulted in only six objects being recognized as noise, compared to 15 for the word2vec and GloVe representations. Among nine identified clusters we can see clearly distinguishable categories such as
\textit{type of vehicle}, \textit{animal}, \textit{natural earth formation}, \textit{kitchen utensil}. 

\begin{table}
\begin{center}
\begin{tabular}{l|cc|cc}
  \multirow{2}{*}{Model} & \multicolumn{2}{c|}{AP} & \multicolumn{2}{c}{HDBSCAN} \\
   & SK & VRC & SK & VRC \\
  \hline
  \textbf{CMPM} & \textbf{0.2346} & \textbf{9.6993} & \textbf{0.1833} & \textbf{6.0923} \\
  \textbf{GloVe} & 0.1108 & 3.8113 & 0.0445 & 2.081 \\
  \textbf{word2vec} & 0.0402 & 2.3022 & 0.035 & 1.879 \\
\end{tabular}
\end{center}
\caption{Silhouette Coefficient (SK) and Variance Ratio Criterion (VRC) of the compared models.}
\label{clust_coef}
\end{table}

To furthermore estimate the clustering validity in a quantitative manner, we computed the Silhouette Coefficient \cite{rousseeuw1987silhouettes} and the Variance Ratio Criterion \cite{calinski1974dendrite} for the clusters yielded by applying both algorithms. As reported in Table \ref{clust_coef} the results of the graph- and density-based cluster analysis performed on our model score higher on the both used coefficients. 
The value of the silhouette coefficient for the word2vec representations is close to zero which indicates overlapping clusters. Also the difference in the SK scores for the GloVe vectors and our model cannot be considered negligible and suggests that clusters identified using GloVe representations contain more misplaced objects. 
The variance ratio criterion is in contrast to the the silhouette coefficient not normalized to a fixed range, making a comparison of different models difficult. However, it is observable that for all models the value of the VRC for the results of HDBSCAN is generally lower than for the AP clustering. Comparing the GloVe representations and our model, we see a stronger decline in the index value for the GloVe vectors, which can imply a more robust clustering structure of our embeddings. 

Based on the conducted qualitative and quantitative evaluation, we can conclude that the clustering patterns obtained from our embeddings outperform the two used baselines indicating dense and better separated clustering structure. 
This, along with the finding stated above that using our representations associated entities are being grouped together, suggests that our embeddings are well-constructed and result in clearer entity categorisation.

\section{Related Work}
\label{Sec:relatedwork}


The word2vec model architecture proposed by Mikolov et al. \cite{mikolov2013efficient} gained great popularity and gave a rise for further research in the domain of neural language modelling. Since its introduction multiple uni-modal modifications of the initial word2vec model \cite{levy2014dependency, cui2014framework, coulmance2016trans} as well as independent frameworks based on other approaches \cite{luong2013better, bojanowski2017enriching, pennington2014glove, upadhyay2017beyond, peters2018deep} were developed. 
Natural language both written and spoken is often accompanied by visual signals in form of images and videos. This observation from the real world further motivated the development of multi-modal representations combining learning word and image vectors \cite{srivastava2012multimodal, frome2013devise, kiros2014multimodal, yan2015deep, collell2017imagined, wang2018learning}.

However, previously proposed models do not consider neither mixing modalities in the initial embedding step nor exploiting text layout and design elements serving as a guide for human perception. Thus, they do not reflect cognitive processes underlying human knowledge acquisition and contradict characteristics of how humans perceive and interact with documents integrating both textual and visual data which can potentially enable them to capture the context better and contribute to further performance enhancement. 
Observers of multimodal documents often shift the focus of their attention between the text and accompanying images integrating their mutually beneficial, complementary characteristics to achieve better comprehension of the media content \cite{holmqvist2005role, bucher2006relevance}. 
The gaze of an observer commonly leaps between pictures, graphics as well as text units, for instance, headlines, drop quotes or fact boxes in the context of newspaper reading \cite{holsanova2006entry}. The viewer tries to determine the relative importance of the individual components of the document based on the visual saliency of the corresponding elements and searches for entry points. The latter are defined as points that attract the initial attention of the observer such that browsing-like behaviour stops and in-depth reading of the belonging data content starts \cite{garcia1991eyes}. 
Hegarty \cite{hegarty1991diagrams} who analyzed eye behaviour of subjects studying technical texts accompanied by diagrams reports similar results to those obtained in our eye-tracking measurements. Subjects interacted with both language and image elements frequently switching the direction of their gaze from reading the text to viewing the corresponding picture. While text was perceived as a key part of the provided multimodal document, the according pieces of the diagram were usually visited after processing the sentence relating to it. 

These observations made in the domain of media research potentially have far reaching implications for machine learning techniques providing additional support of the idea of the integration of multiple media types in the learning models. The transfer of the patterns of the human-like interaction with multimedia content in the training algorithms is a challenging task, since multimodal content usually provides multiple entry points and reading paths, thus being multisequential \cite{holsanova2014reception}. 
As was shown by multiple research studies, the differences in visual behaviour can be influenced by design and layout of the multimodal message \cite{holsanova2005tracing}, medium type \cite{bucher2006relevance}, particular intentions of the recipients of the multimedia content \cite{yarbus2013eye}, but also their individual characteristics such as gender, age, origin, cognitive abilities, interests and prior knowledge. 

\section{Conclusions and Future Work}

While distributed representations, a.k.a~embeddings, are the core component of representation learning, current machine-learning algorithms devised to train them don't exploit all information given in the documents of their source data. They ignore multimodal information and don't consider
layout information. 

In this paper we tested the hypothesis that context models which are inspired by human media perception have advantages over basic models like word windows. The underlying rationale being, that media documents are designed by humans for humans, not for machines and hence are better perceived in a human-like way.

To develop a computational model for parsing multimedia documents that is inspired by how humans interact with multi-modal information we first conducted eye tracking studies on simple documents generated from the Visual Genome data set. The generate documents consist of a headline, an image and a description, plus an underlying graph-structured document representation. Based on the obtained gaze trajectories we derive a basic computational trace model, called CMPM, that allows to compute a sequence of multimodal entities as they might be observed by humans. 

Based on the CMPM we extended the word2vec skip-gram model in order to capture multimodal context and layout elements. We compared the obtained representations to the pre-trained word2vec and GloVe embeddings on a variety of tasks. 
Our findings indicate that the integration of human-like perception patterns into a cross-modal framework provides an additional source of information and has a positive influence on the overall performance. 

Our model encourages future work in this direction since it is kept as simple as possible and makes additional gains with more sophisticated approaches very likely.
An obvious direction of future research is the scaling to larger data sets. To achieve the objective of this work we had to adapt an openly-available multi-modal data collection. 
A large-scale data set designed specifically for research on perceptual models and offering more meaningful text and richer image material as exploited in this study is required to enhance the performance of the proposed model. 

Furthermore, our research was restricted to documents with very basic content and layout. Follow-up studies could expand the application of our cross-modal perception-trace approach on other types of multi-media documents, e.g. information graphics, websites or their extracts such as social media posts or display advertising consisting of both textual and visual units. 


Finally, our study focuses on the basic question whether entity representations can be optimized by exploiting a prototypical model of how humans perceive multi-media data. 
While we found clear positive indications that our approach brings measurable performance gains and, thus, can be a first step towards more human-like perception and cognitive understanding of documents by computers, 
the development of more sophisticated learning algorithms 
promise great prospects for further research.

\bibliographystyle{unsrt}  
\bibliography{sample-base}

\end{document}